\documentclass[a4paper,fleqn]{cas-dc}

\usepackage[numbers]{natbib}

\usepackage{epstopdf}
\usepackage{tikz}
\usepackage{algorithm}
\usepackage{algorithmic}
\usepackage{graphicx}
\usepackage{array}
\usepackage{hyperref}
\usepackage{cleveref}
\usepackage{caption}

\begin{document}
\let\WriteBookmarks\relax
\def\floatpagepagefraction{1}
\def\textpagefraction{.001}
\shorttitle{From implicit to explicit feedback}
\shortauthors{Quyen et al}

\title [mode = title]{From Implicit to Explicit feedback: A deep neural network for modeling sequential behaviours and long-short term preferences of online users}  

\tnotemark[1]
\tnotetext[1]{A part of this work appears in \cite{DBLP:conf/acml/TuanTTLT19}}


\author[1]{Quyen Tran} 
\cormark[1]
\ead{tranquyenhd17@gmail.com}

\author[1]{Lam Tran}
\cormark[1]
\ead{lam.tt173226@gmail.com}

\cortext[cor1]{Equal contribution}

\author[2]{Linh Chu Hai}
\ead{chuhailinh29071998@gmail.com}

\author[1]{Linh Ngo Van}
\ead{linhnv@soict.hust.edu.vn}

\author[1]{Khoat Than}
\ead{khoattq@soict.hust.edu.vn}

\address[1]{Hanoi University of Science and Technology, No. 1, Dai Co Viet road, Hanoi, Vietnam}
\address[2]{VCCorp Corporation, Vietnam}

\begin{abstract}
In this work, we examine the advantages of using multiple types of behaviour in recommendation systems. Intuitively, each user has to do some \textbf{implicit} actions (e.g., click) before making an \textbf{explicit} decision (e.g., purchase). Previous studies showed that implicit and explicit feedback have different roles for a useful recommendation. However, these studies either exploit implicit and explicit behaviour separately or ignore the semantic of sequential interactions between users and items. In addition, we go from the hypothesis that a user's preference at a time is a combination of long-term and short-term interests. In this paper, we propose some Deep Learning architectures. The first one is \textbf{Implicit to Explicit (ITE)}, to exploit users' interests through the sequence of their actions. And two versions of ITE with Bidirectional Encoder Representations from Transformers based (BERT-based) architecture called \textbf{BERT-ITE} and \textbf{BERT-ITE-Si}, which combine  users' long- and short-term preferences without and with side information to enhance user representation. The experimental results show that our models outperform previous state-of-the-art ones and also demonstrate our views on the effectiveness of exploiting the implicit to explicit order as well as combining long- and short-term preferences in two large scale datasets. 
\end{abstract}
\begin{keywords}
\sep Deep Learning \sep Recommendation systems \sep Collaborative Filtering \sep Implicit Feedback \sep Explicit Feedback   \sep Long-term preference \sep Short-term preference 
\end{keywords}

\maketitle

\section{Introduction}
\label{sec:1}

Recommendation system (RS) is "the soul" of e-commerce systems in helping to suggest items that match the user's interests. Collaborative Filtering (CF) is the most commonly used approach to build a personalized recommendation system. CF-based methods exploit interaction information between users and items, in the form of implicit \cite{DBLP_HuKV08}, \cite{DBLP:series/sbcs/SymeonidisZ16} (click, view, search, etc.) or explicit feedback \cite{DBLP:conf/nips/SalakhutdinovM07}, \cite{DBLP:conf/cikm/ZigorisZ06} (like, rate, add to cart, buy, etc.) to learn user and item representations for making suggestion.

Regarding the exploitation of different types of behaviours on one item, we believe it would be beneficial to model the order among them. This is because naturally, the chain of actions of a user follows the rule: an explicit action on an item will likely be made if several interactions with that item have been made implicitly before. However, existing works that take advantage of both these two types either only consider them separately or use a simple method to model such order \cite{DBLP:conf/cikm/LiuXZY10}, \cite{koren2011recommender}, \cite{DBLP:conf/cscwd/ShiLLZG17}, \cite{DBLP:conf/icde/Gao0GCFLCJ19}, \cite{li2016exploiting}. For example, Multi-task Matrix Factorization (MTMF) \cite{DBLP:conf/cscwd/ShiLLZG17} focuses on predicting user's product rating (explicit feedback) with the help of data about implicit behaviour (view, want). Nevertheless, these behaviours are exploited separately before being combined to predict the rating, which may fail to capture their ordinal relation. Unlike MTMF, Neural Multi-task Recommendation (NMTR) \cite{DBLP:conf/icde/Gao0GCFLCJ19} tries to build a model that can describe the order of these types of behaviours. The behaviours are sorted in the ascending order of "degree of clarity": the
likelihood of a type of behaviours is determined depending on the likelihood of the previous type. However, this sequential impact is simply encoded by a scalar addition, which might not be effective enough. We instead aim to better capture the sequence of behaviours from implicit to explicit by using a more complex architecture and modeling them in two consecutive phases, in single-task instead of multi-task.

In another aspect, modeling comprehensively user's preferences is likely to achieve better performance. Typically, user's preferences can be divided into two kinds: Short-term and long-term preferences. Short-term preference of a user reflects his/her interest in a short period of time, which can be discovered in recent items that user has interacted with. In contrast, long-term preference shows aspects that are more stable in user's interest, or in other words, it shows what the user truly likes. This kind of preference usually appears along with the whole interaction history of the user. 

In order to model user's short-term preference, it is necessary to exploit the sequence of recently interacted items. This could be done by adopting sequential models in Natural Language Processing (NLP) due to the similar characteristics between item sequence and words. For instance, Long- and Short-term User Representations (LSTUR) \cite{DBLP:conf/acl/AnWWZLX19} uses the architecture of Gated Recurrent Unit (GRU) \cite{DBLP:conf/emnlp/ChoMGBBSB14} to capture the relationship among the news in the user history; or some models such as Sequential Recommendation with BERT (BERT4Rec) \cite{DBLP:conf/cikm/SunLWPLOJ19}, Self-Attentive Sequential Recommendation (SASRec) \cite{DBLP:conf/icdm/KangM18} employ the self-attention mechanism to better model the sequence of recent items. Existing recommendation methods following this item sequence exploitation scheme can be divided into two groups: one is session-based recommendation \cite{DBLP:conf/cikm/SunLWPLOJ19, DBLP:conf/icdm/KangM18} and one is personalized CF \cite{DBLP:conf/acl/AnWWZLX19}. The former group aim to find the next item of the current session, but cannot be applied in personalized CF because it does not learn user representation. The latter group 
exploit the sequence of recently interacted items as a supplement for learning user embedding. However, to the best of our knowledge, they ignore the different types of behaviour on an item. Meanwhile, our work, belonging to personalized CF, aims to model not only user's temporal preference from the recent items, but also the implicit and explicit behaviours on an item in two consecutive phases.

In terms of modeling user's long-term preference, pure CF-based models can attain to some extent since they learn from the whole user-item matrix. However, it could still be improved if additional information related to user's long-term interests is utilized. Such additional information can be the content of items (\cite{DBLP:conf/mm/HuangQFSX18}, \cite{DBLP:conf/kdd/WangWY15}), or simply the embedding of user identification (user ID) \cite{DBLP:conf/acl/AnWWZLX19}. In our work, we attempt to incorporate the high-level aspect of long-term preference in the form of item category into learning user representation. More specifically, we base on the intuition that if a user likes a certain category, he/she will likely interact with more items belonging to that category. In other words, the frequency of categories throughout the interaction history of one user can reflect some information about his/her long-term preference. Moreover, we choose this type of side information because of its easy-to-collect attribute in real-world systems, compared to other types such as the content of items which requires more computational resource to process and might contain noises. 


To sum up, our work has the following contributions:

\begin{itemize}
    \item We propose model ITE to exploit the implicit-to-explicit order of user behaviour on an item. Accordingly, our model includes two main modules: Implicit and Explicit modules. Basically, ITE learns the implicit interactions between user and corresponding target item through Neural Matrix Factorization (NeuMF) \cite{DBLP:conf/www/HeLZNHC17} architecture, to get a representation vector at the end layer of the Implicit module. And this vector will be the input of the Explicit module. Therefore this model helps to discover the complex relationship between implicit and explicit behaviours.
    
    \item Additionally, we enrich user preference information by combining long-term interest information with short-term interest information through a BERT-based architecture called BERT-ITE. Instead of using NeuMF as in ITE, BERT-ITE uses a BERT-based module with the input vectors are embedding vectors of user and corresponding items that he/she interacted with within a recent session. This BERT-based module learns the relationship in terms of context of these vectors, thus it helps to exploit user's preferences in a complete way.
    
    \item Also, we enhance user's long- and short-term preferences information by using additional side information in BERT-ITE-Si (BERT-ITE's extended version), based on category information, on user and item original representations. On the one hand, it helps to represent items more efficiently because this specific information is rich in the meaning of each item (which categories the item belongs to). On the other hand, side information carries information about user's long-term interests (user engagement rate w.r.t categories). As mentioned above, in BERT-ITE model, we exploit the user's short-term preferences through the item sequence that user has recently interacted with. So, in this way, we can improve user representation and better capture user's long- and short-term preferences efficiently. 
\end{itemize}

We also conducted extensive experiments on two large scale datasets. The results show that our models have outstanding performances compared to previous models. In addition, once again side information is proved to be a useful resource when the results of the models attached to external data are significantly improved.

The rest of this paper will be laid out as follows: Section \ref{sec:related} and \ref{sec:background} will briefly review some related work and background knowledge. Section \ref{sec:pro_model} will detail our proposed models and discuss some of their advantages. Section \ref{sec:experiment} will present scenarios and results of experiments of proposed methods, compared with some previous methods. Finally, we conclude in Section \ref{sec:conclusion}.

\section{Related work}
\label{sec:related}

In this section, we will briefly review some studies related to ours. We discuss general recommendation, followed by experienced-based and transaction-based recommendations. General recommendation only uses one type of user-item behaviour interactions, i.e. either implicit feedback or explicit feedback. Meanwhile, the term experienced-based recommendation refers to exploiting multiple behaviours on an item and transaction-based recommendation refers to considering a sequence of items. These two term are adopted from \cite{DBLP:journals/tois/FangZSG20}, and we suggest readers visit that paper for a broader view about sequential recommendation.  
	
In general recommendation, modeling the interaction of users and items based on users’ historical feedback has been the main focus of many approaches. This could be done by finding the latent representation of user and item in a mutual space then taking their inner product \cite{DBLP:reference/sp/2015rsh}, \cite{DBLP:reference/sp/KorenB15}. In order to capture the complex combination of these latent vectors, which the conventional Matrix Factorization (MF) could not handle, Xiangnan He et al. proposed Neural Matrix Factorization (NeuMF) \cite{DBLP:conf/www/HeLZNHC17}. This model takes advantage of the ability of deep learning architecture when finding complex patterns. In terms of data, due to the sparsity of explicit feedback, many CF-based approaches only consider implicit one, which is in a form of binary matrix, and use different motivations and strategies to utilize it. For example, Pair-wise Probabilistic Matrix Factorization (PPMF) \cite{DBLP:journals/ijon/LiO16} and Dynamic Bayesian Logistic Matrix Factorization (DBLMF) \cite{DBLP:conf/ijcai/LiuZLLGLJ18} focus on enriching user and item representations by respectively applying a pair-wise ranking algorithm to probabilistic matrix factorization and modeling their heterogeneous evolvement with a diffusion process at each timestamp. 
Another solution to the problem of sparse explicit data is to reasonably infer implicit information from the explicit. Take Recommendation with multiplex implicit feedbacks (RMIF) \cite{DBLP:journals/ijon/HuXLWXC20} as an example, three types of implicit feedback are heuristically generated based on user similarities, item's rated records and user's positive attitudes. Then learnable multiplex implicit vectors are created and served as auxiliary vectors for user/item's latent vectors when predicting user's rating for an item. To sum up, learning the user-item interaction only relying on one interaction matrix may not be optimal. This is the motivation for us to consider more types of interactions between user and item as well as other aspects that could enrich user representation. 

	
Multi-behaviour recommendation or experience-based recommendation, which considers both users’ implicit and explicit feedback, has recently been paid more attention due to the effectiveness of modeling the sequence of behaviours on an item. Studies of this class can be divided into two groups. One group takes both implicit and explicit feedback into consideration, however, separately \cite{DBLP:conf/kdd/SinghG08, DBLP:conf/wsdm/Krohn-GrimbergheDFS12, DBLP:conf/www/ZhaoCHC15, li2016exploiting, ijcai2018-509, DBLP:journals/apin/MandalM20, DBLP:conf/cscwd/ShiLLZG17}. For example, MTMF \cite{DBLP:conf/cscwd/ShiLLZG17} treats ‘views’ and ‘wants’ feedback as compliments for better rating prediction and uses a common user feature space to factorize matrices of all types of feedback. The other group aims to model the ordinal relation between them. The typical example of this group is NMTR \cite{DBLP:conf/icde/Gao0GCFLCJ19}. It manages to model the cascading relationship among multiple types of behaviour and adopts the multi-task learning framework to learn the model in which each task is a behaviour. However, only adding the prediction of the previous behaviour (a scalar) to the current one to encode the behaviour sequence is too simple, thus could not fully capture the complex semantic relationship among various behaviour types. In our work, we aim to model the nature of this behaviour sequence in a more complex and reasonable way.  

In terms of transaction-based recommendation, in addition to methods that use Markov chains (MCs) (\cite{DBLP:conf/recsys/HeKM17}, \cite{DBLP:conf/icdm/HeM16}, \cite{DBLP:conf/www/RendleFS10}, \cite{DBLP:journals/jmlr/ShaniHB05}), Recurrent Neural Network based (RNN-based) models have been well exploited (\cite{DBLP:conf/recsys/DonkersL017}, \cite{DBLP:conf/cikm/HidasiK18}, \cite{DBLP:journals/corr/HidasiKBT15}, \cite{DBLP:conf/cikm/LiRCRLM17}, \cite{DBLP:conf/recsys/QuadranaKHC17}, \cite{DBLP:conf/wsdm/WuABSJ17}, \cite{DBLP:conf/sigir/YuLWWT16}) to capture sequential patterns. The main idea of these models is to learn a vector representing all items in a sequence to predict the next item like in Session-based recommendation, or to model user's short-term preference when considering user representation. For example, session-based Recommendation with RNN (GRU4Rec) \cite{DBLP:journals/corr/HidasiKBT15} uses Gated Recurrent Units (GRU) \cite{DBLP:conf/emnlp/ChoMGBBSB14} with each item going into one unit to model click sequence. Other models were proposed to enhance GRU4Rec \cite{DBLP:journals/corr/HidasiKBT15} in terms of performance, such as sampling strategies \cite{DBLP:conf/iclr/LiuSPGSKS18}, etc. Furthermore, using Convolutional Neural Network (CNN) can alleviate some limits of RNN-based models, such as expensive training costs, to a certain extent, and thus CNN-based models have been proposed. For example, Caser \cite{DBLP:conf/wsdm/TangW18} applies convolution operation horizontally and vertically on the embedding matrix of previous items. Recently, with the promising potential in NLP tasks, attention mechanism has been employed in almost all methods that deal with sequential behaviours (Bert4rec \cite{DBLP:conf/cikm/SunLWPLOJ19}, Deep Feedback Network (DFN) \cite{DBLP:conf/ijcai/XieLWWXL20}, Deep Knowledge-Aware Network (DKN) \cite{DBLP:conf/www/WangZXG18}, News recommendation with Attentive Multi-view Learning (NAML)  \cite{DBLP:conf/ijcai/WuWAHHX19}, HyperNews \cite{DBLP:conf/ijcai/LiuPCZ20}, LSTUR \cite{DBLP:conf/acl/AnWWZLX19}, News recommendation with Multi-head Self-attention (NRMS) \cite{DBLP:conf/emnlp/WuWGQHX19}, News recommendation with Personalized Attention (NPA) \cite{DBLP:conf/kdd/WuWAHHX19}, SASRec \cite{DBLP:conf/icdm/KangM18}, etc.). For example, DKN \cite{DBLP:conf/www/WangZXG18} uses vanilla attention module to learn user representation by aggregating the representations of their history browsers which are extracted with the help of CNN and Knowledge graph; NAML \cite{DBLP:conf/ijcai/WuWAHHX19} applies attention at word-level and view-level to select informative words and views for news representation, then again uses attention to learn user representation by focusing on important news; Similarly, NPA \cite{DBLP:conf/kdd/WuWAHHX19} uses personalized attention at word- and news-level to learn different preferences of different users. In addition to vanilla attention, self-attention mechanism, which is a core concept of Transformer-based architectures, has been well exploited. For instance, NRMS \cite{DBLP:conf/emnlp/WuWGQHX19} encodes news from title by using multi-head self-attention to model words interaction and similarly encodes user from his/her browsed news. It also applies additive attention on the result of multi-head self-attention to select important words and news for news and user representation respectively. SASRec \cite{DBLP:conf/icdm/KangM18} also successfully exploits this mechanism to balance short-term intent and long-term preference when predicting the item in the item sequence. Bert4Rec \cite{DBLP:conf/cikm/SunLWPLOJ19} improves SASRec \cite{DBLP:conf/icdm/KangM18} by introducing a bidirectional model to capture context from both directions when modeling item sequence. However, one common limit of these approaches is that they only consider the sequence of historical interacted items (thus one type of user feedback is used) to predict the next item or to model user's preferences. Instead, our work aims to further improve the prediction effectiveness by incorporating the behaviour sequence on one item.
	

\section{Background}
\label{sec:background}
In this section, we introduce some preliminaries serving as the base knowledge of our models.


\subsection{Neural matrix Factorization}

\begin{figure*}[pos=!ht]
    \centering\includegraphics[scale=0.5]{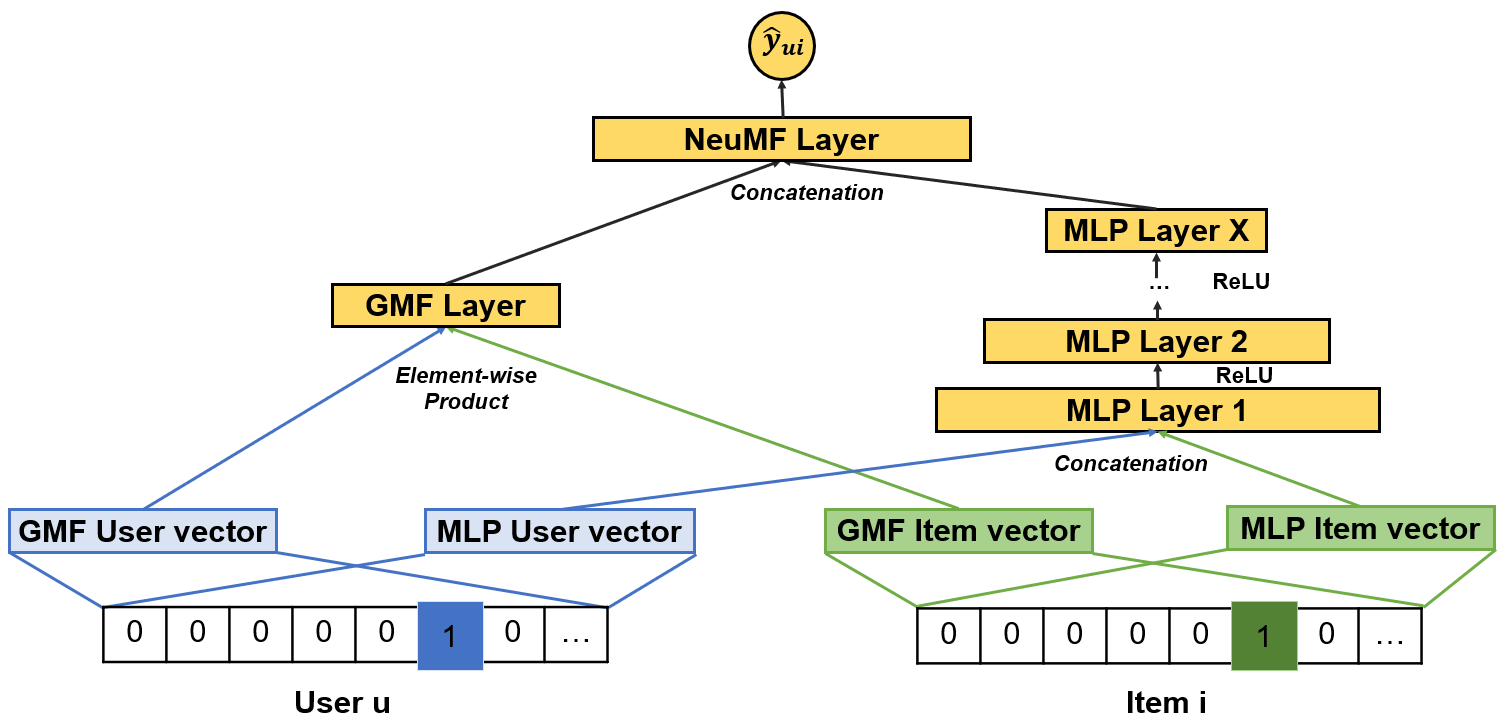}
    \caption{The architecture of NeuMF model}
    \label{fig-neumf}
\end{figure*}

Neural matrix factorization (NeuMF) \cite{DBLP:conf/www/HeLZNHC17} is a fusion mo- del of Generalized Matrix Factorization (GMF) and Multi-Layer Perceptron (MLP) - two instantiations of Neural Collaborative Filtering (NCF), that can learn the interaction between user and item in both linear and non-linear manners. In particular, GMF extends MF by projecting the element-wise product vector of user and item latent representation to the output layer with learnable weights and an activation function. This architecture enables the model to dynamically learn the non-linear interaction from the latent representation instead of treating all dimensions the same (taking sum as in MF). The other version, MLP, first concatenates the user and item latent vectors then passes the concatenated vector through several hidden layers. This provides a more flexible way to model the user-item interaction than GMF which considers the element-wise product. To utilize the advantages of the two models, the authors let each of them learn their latent embeddings separately before concatenating the two final layers and projecting it to the output layer. The architecture of NeuMF is presented in Fig. \ref{fig-neumf}.

\subsection{BERT and BERT4Rec}
\label{subsec:bert4rec}
BERT \cite{DBLP:journals/corr/abs-1810-04805} is a language model that is trained bidirectionally. In other words, it looks at a text sequence from both left-to-right and right-to-left directions, thus, the context of a word could be better captured. It consists of several layers stacking on top of one another where each layer is built upon the self-attention layer - Transformer layer \cite{DBLP:conf/nips/VaswaniSPUJGKP17}. 

\begin{figure*}[pos=!ht]
    \centering\includegraphics[scale=0.39]{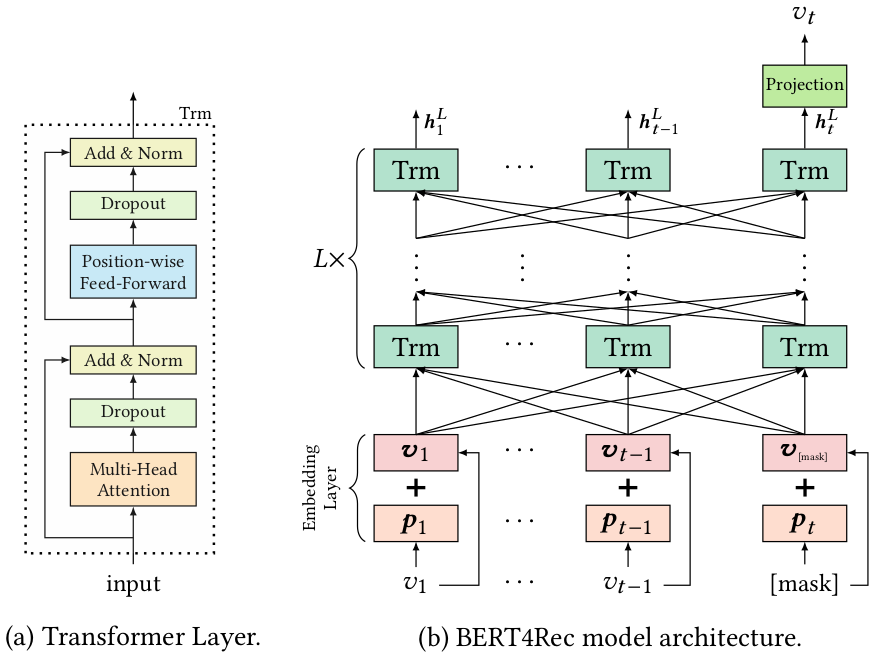}
    \caption{BERT4Rec architecture \cite{DBLP:conf/cikm/SunLWPLOJ19}}
    \label{fig-bert4rec}
\end{figure*}

Fig. \ref{fig-bert4rec}b demonstrates architecture of BERT4Rec \cite{DBLP:conf/cikm/SunLWPLOJ19}. It is composed of an embedding layer and $L$ Transformer layers. It receives an item sequence of length $t$ as input and computes the final hidden vector representation of each item (the output of the $L^{th}$ layer). In the BERT4Rec paper \cite{DBLP:conf/cikm/SunLWPLOJ19}, these hidden vector representations are then fed into a two-layer feed-forward network to predict the masked item. However, we only utilize the BERT architecture to learn the representation of user from item sequence, thus, we will explain in detail the architecture up to this part. As shown in Fig. \ref{fig-bert4rec}a, one Transformer layer consists of two sub-layers, a Multi-Head Self-Attention sub-layer and a Position-wise Feed-Forward sub-layer. Note that the superscript $l$ is used to represent the $l^{th}$ layer.

\textbf{Transformer layer}

\emph{Multi-Head Self-Attention - $MH$} 




Assume $H_l$ is the input matrix of the $l^{th}$ layer (each row representing each item); $h$ is the number of heads. Each head is a Scaled Dot-Product Attention \cite{DBLP:conf/nips/VaswaniSPUJGKP17}:
\begin{equation}
    \begin{aligned}
        &head_i = Attention(H^l W_i^Q, H^l W_i^K, H^l W_i^V), \\
        \text{where } &Attention(Q, K, V) = softmax(\frac{QK^T}{\sqrt{d/h}})V 
    \end{aligned}
\end{equation}
For each $head_i$, $W_{qi}, W_{ki}, W_{vi}$ $\in$ $R^{d\times d/h}$ are the learnable matrices projecting $H^l$ into different spaces to compute its self-attention vector. Then these $h$ heads are concatenated and projected once more time to get the Multi-Head Self-Attention:
\begin{equation}
    \begin{aligned}
        &MH(H^l) = [head_1; head_2;... head_h]W^O, \\
        \text{where } &W_o \in R^{d\times d}
    \end{aligned}
\end{equation}

\emph{Position-wise Feed-Forward Network - $PFFN$} 

Because the self-attention ultimately is a linear model, a position-wise feed-forward network is applied identically on every row of $MH(H^l)$ to bring nonlinearity to the model and enable interactions between different latent dimensions. For simplicity, $MH(H^l)$ is denoted as $A$:
\begin{equation}
    \begin{aligned}
        &PFFN(A) = [FFN(A_1)^T, ..., FFN(A_t)^T]^T, \\
        &FFN(x) = GELU(xW^{(1)} + b^{(1)})W^{(2)} + b^{(2)}, \\
        &GELU(x) = x\Phi(x).
    \end{aligned}
\end{equation}
where $\Phi(x)$ is the cumulative distribution function of the standard Gaussian distribution; $W^{(1)} \in R^{d\times 4d}$, $W^{(2)} \in R^{4d \times d}$, $b^{(1)} \in R^{4d}$, $b^{(2)} \in R^d$ are weight matrices and biases of the two-layer feed-forward network.

\textbf{Stacking Transformer Layer} 

In order to learn more complex item transitions, several transformer layers are stacked. In addition, dropout \cite{DBLP:journals/jmlr/SrivastavaHKSS14}, layer normalization \cite{DBLP:journals/corr/BaKH16} and residual connection \cite{DBLP:conf/cvpr/HeZRS16} are employed to face some problems of deep network \cite{DBLP:conf/icdm/KangM18}. The operations are the same for each sublayer, and are illustrated in figure 2a).

\section{Proposed models}
\label{sec:pro_model}
In this section, we first present our proposed models, including ITE, BERT-ITE and BERT-ITE-Si. We will then discuss some advantages of these models: the order from implicit to explicit feedback, the exploitation of short-term in addition to long-term preferences, as well as our enhancement of these two types of preference by using side information.

\subsection{Notations}
\label{subsec:notation}
 First, we clarify the data used in the class of ITE models. As mentioned before, our models aim at modeling the sequence of multi-behaviour. Therefore, the models utilize multiple types of user-item interactions, including implicit and explicit feedback. In this paper, we define implicit behaviour is the action showing that a user wants to know more about an item such as \textbf{view}, \textbf{click}. In contrast, explicit behaviour is the action showing that a user is fascinated with an item such as \textbf{purchase}, \textbf{order}, \textbf{add to cart}. A user tends to like an item before making a purchase. Intuitively, a user may perform some implicit behaviours before doing an explicit action. Another type of data used in this paper is the side information. Side information often exists as item features, movie categories, music genres, etc. and can be encoded as a T-dimensional vector. 

Second, we provide general encoding rules for user and item representations and their interaction in this paper as follows:
\begin{itemize}
    \item $M, N$: are the number of users and items respectively.
    \item $K$: is the dimension of the user or item embedding vectors transformed from original representations.
    \item $\mathbf{u}, \mathbf{i}$: are the representation vectors of user u and item i respectively, where $\mathbf{u} = (\alpha_{1}, ..., \alpha_{S}), \mathbf{i} = (\beta_{1}, ..., \beta_{W})$.
    \item $X = (x_{ui})_{M\times N}$: is the implicit data, where $x_{ui}$ indicates whether user $u$ interacts with item $i$ or not.
    \item $Y = (y_{ui})_{M\times N}$:  is the explicit data, where $y_{ui}$ represents user $u$'s interest in item $i$.
\end{itemize}

In addition, our models can employ side information to enrich the representations. In some situations, the one-hot encoding suffers from the lack of information, thus it is expected that side information can enhance their representations. Additionally, side information is expected to help the models avoid cold start problem as mentioned in much previous research \cite{DBLP:journals/ijar/LeCTLT18, DBLP:conf/rivf/HoangTLT19}. In this case, we denote the following:
\begin{itemize}
    \item $T$: is the number of dimensions of the vector representing side information.
    \item $u^{(m)}, i^{(m)}$: are the same as the original vectors $\mathbf{u}, \mathbf{i}$ of user and item respectively.
    \item $u^{(a)}, i^{(a)}$: are the vectors constructed from \textit{item category} data. More specifically, $ i^{(a)} = (\theta_1, \theta_2, ..., \theta_T)$ denotes the category information of item $i$. Then, $u^{(a)} = (\varphi_1, \varphi_2, ..., \varphi_T) = (\frac{\epsilon_1}{\epsilon}, \frac{\epsilon_2}{\epsilon}, ...,\frac{\epsilon_T}{\epsilon})$, $\epsilon_j$ denotes the number of items interacted by user $u$ belonging to the $j^{th}$ category, $\epsilon = \epsilon_1 + \epsilon_2 + ... + \epsilon_T$ and $\varphi_j = \frac{\epsilon_j}{\epsilon}$. 
\end{itemize}

\subsection{Implicit to Explicit (ITE) Model}
\label{subsec:ITE}

\begin{figure*}[pos=!ht]
    \centering\includegraphics[scale=0.5]{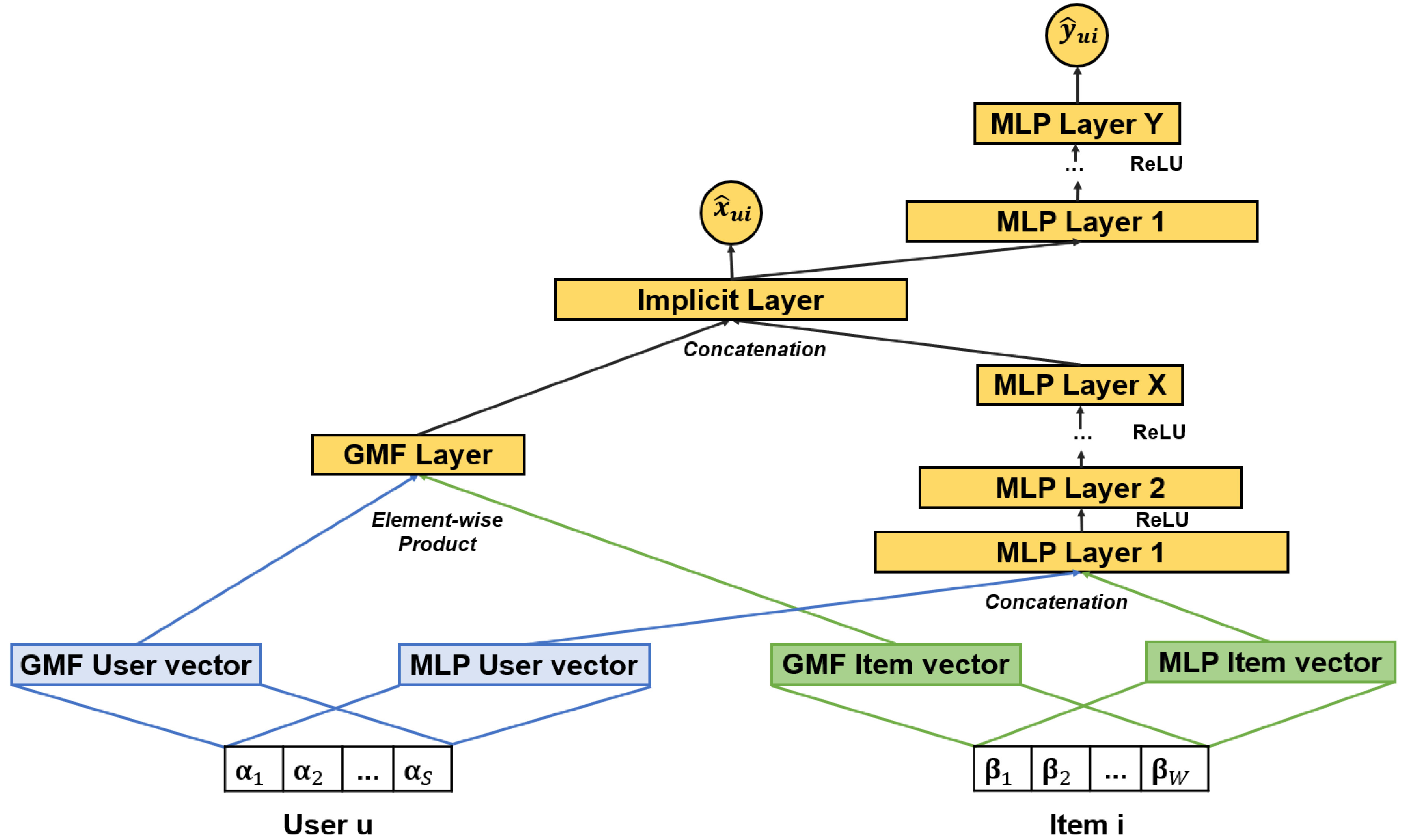}
    \caption{Implicit to Explicit (ITE) architecture}
    \label{fig-ite}
\end{figure*}

We present the \textbf{Implicit to Explicit (ITE)} architecture in Fig.~\ref{fig-ite}. The input layer is the composition of user $u$ and item $i$ representations. These representation can be adjusted depending on available data, such as one-hot encoding (Koren et al \cite{DBLP:journals/computer/KorenBV09}, Gao et al \cite{DBLP:conf/icde/Gao0GCFLCJ19}), neighbor-based encoding (Rendle \cite{DBLP:conf/icdm/Rendle10}). In this work, we use one-hot encoding to represent the inputs.

The input vectors is then fed into a neural network which represents the \textbf{implicit module} of the model. We utilize the NeuMF architecture~\cite{DBLP:conf/www/HeLZNHC17} for this module due to its good performance when dealing with implicit data. The last hidden layer of the \textbf{implicit module} is called \textbf{Implicit Layer}. Let $p_u^G, q_i^G$ denote the embedding vectors of user $u$ and item $i$ for GMF layer. $p_u^M, q_i^M$ is similarly denoted for MLP Layer 1. Let $\phi^{GMF}, \phi^{MLP}$ be the output vectors of GMF Layer and MLP Layer X shown in Fig.~\ref{fig-ite}. We will formulate the implicit module as what follows.

The hidden layer of GMF part can be expressed as follows:
\begin{equation}
    \phi^{GMF} = p_u^G \odot q_i^G
\end{equation}
where $\odot$ denotes the pair-wise product. MLP user vector and MLP item vector are fed into a Multi-Layer Perceptron network including X hidden layers. Particularly, each layer performs the following computation:
\begin{equation}
    \phi_{(l+1)}^{MLP} = f(W_{(l)} \phi_{(l)}^{MLP} + b_{(l)})  
\end{equation}
where $l$ is the layer number, $f$ is the activation function (usually ReLU function), $W_{(l)}, b_{(l)}$ are weight and bias of the $l^{th}$ MLP layer. The GMF layer and MLP Layer X in Fig. \ref{fig-ite} are then concatenated to get the Implicit Layer:
\begin{equation}
    \phi^{I} = \begin{bmatrix} \phi^{GMF} \\ \phi^{MLP}_{(X)} \end{bmatrix}
\end{equation}
where $\phi^I$ is the vector denoting \textit{implicit layer}. From the implicit layer, we can output the likelihood that user $u$ will perform an implicit action on item $i$:
\begin{equation}
    \hat{x}_{ui} = \sigma ( \mathbf{h}_{I}^T \phi^{I})
    \text{, where $\sigma$ is the sigmoid function}
\end{equation}

The sequence of behaviour will be modeled by passing the Implicit Layer to a Multi-Layer Perceptron (MLP) network called \textbf{explicit module}. After several hidden layers of this MLP, we achieve the MLP Layer Y as shown in Fig. \ref{fig-ite}. Reminding that the vector of implicit layer is $\phi^{I}$. The \textit{MLP Layer 1} of explicit module is computed as follows:
\begin{equation}
    \phi^{E}_{1} = f(V_{(0)} \phi^{I} + b_{(0)})  
\end{equation}
Then, each hidden layer of explicit module performs the following computation:
\begin{equation}
    \phi^{E}_{(l+1)} = f(V_{(l)} \phi^E_{(l)} + b_{(l)})  
\end{equation}
where $\textbf{V}, b$ denote weight and bias of hidden layer respectively. Let \textbf{h} denote the edge weight of output layer. The last layer of explicit module $\phi^{E}_{(Y)}$ is used to compute the likelihood that user $u$ will perform an explicit action on item $i$:
\begin{equation}
    \hat{y}_{ui} = \sigma ( \mathbf{h}_{E}^T \phi^{E}_{(Y)})
\end{equation}

Note that the outputs of ITE are $\hat{x}_{ui}$ and $\hat{y}_{ui}$ representing the predicted outcome of implicit and explicit behaviour respectively. In an interpretable way,  $\hat{y}_{ui}$ denotes the probability of user $u$ interacting explicitly with item $i$ given that the implicit behaviour has happened. After learning the model, the probability that user $u$ will interact with item $i$ is expressed by an additional variable $\hat{r}_{ui}$ which strengthens the association between $\hat{x}_{ui}$ and $\hat{y}_{ui}$:
\begin{equation}
    \begin{aligned}
        \hat{r}_{ui} &= \hat{x}_{ui}\hat{y}_{ui}
    \end{aligned}
\end{equation}

\textbf{Architecture note:} As described above, our ITE model has two MLP modules. We choose to adopt the tower pattern design for these two modules from \cite{DBLP:conf/www/HeLZNHC17}, i.e, the lower layer has twice as many nodes as the next layer, as this is a familiar design for MLP in many deep learning models. 



\begin{figure*}[pos=!ht]
    \centering\includegraphics[scale=0.85]{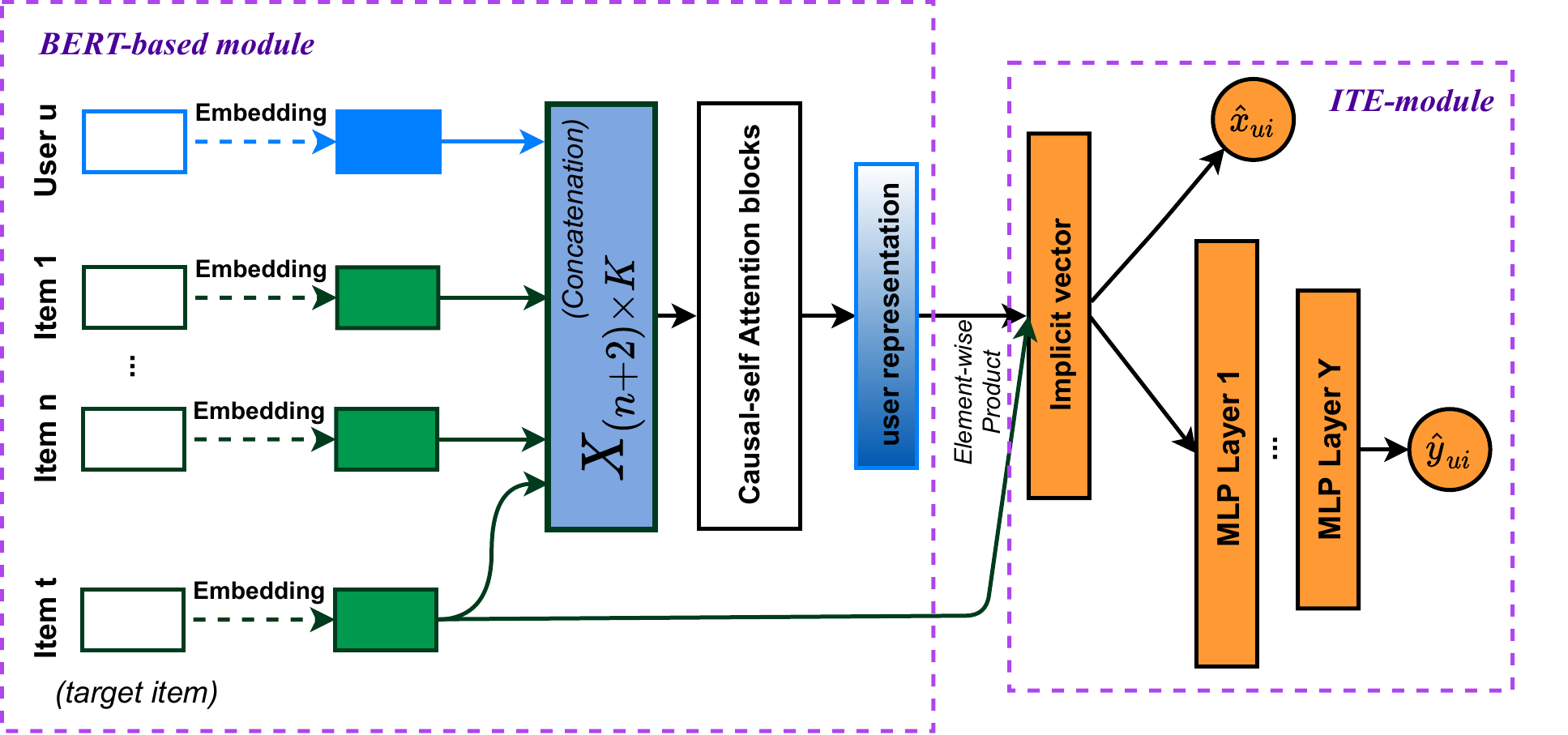}
    \caption{The BERT-ITE architecture}
    \label{fig-md1}
\end{figure*}

\subsection{BERT-ITE Model}
\label{pro-bertIte}


In this subsection, we proposed an extended version of ITE, called BERT-ITE. This model architecture is demonstrated in Fig.~\ref{fig-md1}. The model consists of two main parts, which are also two modules that exploit two important relationships: the BERT-based module inspired by BERT4Rec \cite{DBLP:conf/cikm/SunLWPLOJ19} exploits the relationship between user representation and representations of items which are interacted in that user's recent session, and the ITE-module exploits the order of explicit and implicit interaction of a user with a particular item as mentioned above in ITE models. It is easy to see that: compared with the original ITE model, BERT-ITE model uses BERT-based module instead of NeuMF module. 


The input vectors respectively represent user and items in historical order of session (the last item is the target item), then concatenated into $X$ matrix of size $[n + 2, K]$ and as input to the BERT-based architecture. Where $n$ is length of item sequence before target item, K is embedding dimension number of user or item vectors transformed from original ones. Firstly, $X$ matrix is passed through $L$ Transformer layers \textit{(Trm)} as described in \ref{subsec:bert4rec}:
\begin{equation}
    \begin{aligned}
        &X^l = Trm(X^{l-1}), 	\forall l \in [1,...,L] \\
        &Trm(X^{l-1}) = LN(A^{l-1} + Dropout(PFFN(A^{l-1}))), \\
        &A^{l-1} = LN(X^{l-1} + Dropout(MH(X^{l-1}))),
    \end{aligned}
\end{equation}

where $X^l$ is input of the \textit{l-th} Transformer layer and $LN$ is the Layer Normalization operation. Therefore, we have: $X = X^1$ with $l = 1$ and $X^L$ is the output of the last layer from which we can have the user representation vector: 


\begin{align}
    u_{rep} = X^L[0]
\end{align}

The vector $u_{rep}$ is the representation of the user after learning the relationship with the sequence of items in the session. This vector will be element-wise multiplied with the embedding vector $i_{target}^{'} $ of the item $i_{target}$. The obtained result is implicit vector $\phi_I$. This implicit vector will be the input to the next module - ITE module that models the relation between the implicit and explicit behaviour of user u on item $i_{target}$.
\begin{equation}
    \phi_I = u_{rep} \odot i_{target}^{'}
\end{equation}

where $\odot$ is the symbol of the element-wise multiplication. In the ITE module, this $\phi_I$ is both used to predict implicit behaviour and as input to explicit module.

The working mechanism of this ITE module is shown in the following formula:
\begin{equation}
    \begin{aligned}
        &\hat{x}_{ui} = \sigma(h_I^T \phi_I), \\
        &\hat{y}_{ui} = f^{E}(\phi_I)
    \end{aligned}
\end{equation}

With $\phi_I$, we calculate the possibility that $\hat{x}_{ui}$ indicates user u will perform an implicit behaviour on item $i_{target}$. Moreover, the behaviour sequence will be modeled by passing $\phi_I$ through a MLP network $f^E$ - also known as the explicit module as mentioned above. The output of this module is $\hat{y}_{ui}$.

Again, similar to ITE, $\hat{y}_{ui}$ is usually used as the output predicting the confidence that the user will perform an explicit behaviour on the target item. However, to better denote the ordering of the implicit and explicit behaviour - that is, chances user u will make an explicit interaction on item $i_{target}$ having known some implicit behaviours.  After learning the model, the final predicted score is determined as:
\begin{equation}
    \hat{r}_{ui} = \hat{x}_{ui}\hat{y}_{ui}
\end{equation}

There are some key differences that should be noted when comparing our BERT-based module with the architecture of BERT4Rec. First, only the embeddings of items in user history are taken as input in BERT4Rec. On the other hand, we input both user and item embeddings to the BERT-based module. Second, BERT4Rec aims to predict the masked items from the item sequence, therefore, the corresponding outputs of their BERT module are then fed to a feed forward network. However, the purpose of our BERT-based module is to enrich the representation of user by learning the relationship between user and recently interacted items. As a result, only the output corresponding to user is taken to feed to the ITE-module. Third, since we have user embedding in the input of the BERT-based module, we do not use positional embedding. 

In addition, we would like to present in more detail the role of this BERT-based module. This module helps to learn the contextual relationship between user embedding vector and embedding vectors of the items in the sequence (including $i_{target}$). The vector corresponding to the user contains specific information about the user, which can be seen as the information about long-term interest. The corresponding vectors of the items represent information about the items that user has recently interacted with, .i.e short-term interests. Thus, through this BERT-based module, the obtained $u_{rep}$ vector is the vector containing more meaningful information to represent user's interests, by combining information about his or her long- and short-term preferences. Therefore, thanks to the integration of more information about short-term preferences, the user representation has been improved.

\textbf{Architecture note:} Our BERT-ITE model has two main modules: BERT-based module and ITE-module. The MLP layers in the latter module have the same architecture as those in ITE's explicit module, which means they follow the tower pattern. The other one, BERT-based module, consists of several Transformer layers stacking up on each other which in turn includes a multi-head self attention layer. Therefore there are two hyper-parameters that need to be chosen, number of Transformer layers and number of heads. Considering NLP scenarios, long text data usually has much complex semantic information/relation among words and tokens, which requires deeper self attention architecture and more 'heads' to capture. However, in our case, the number of items in sequence as well as the dimensionality of latent vector are relative small compared to those in text sequence, thus using shallower network may already capture the dependencies between items \cite{DBLP:conf/cikm/SunLWPLOJ19, DBLP:conf/icdm/KangM18}. We also empirically evaluate this opinion in the next section with the number of Transformer layers varies.

\subsection{BERT-ITE-Si Model}
In this subsection, we proposed a model that improving user's long- and short-term preferences with side information, based on BERT-ITE model, called BERT-ITE-Si. Our BERT-ITE-Si is the same as BERT-ITE, except that we use additional side information in user and item representations. 



We encode side information directly into the initial representation of the user and items (before putting into the BERT-based module). Compared with BERT-ITE, in BERT-ITE-Si, we append each side information encoding: $u^{(a)}, i^{(a)}$ after the one-hot encoding: $u^{(m)}, i^{(m)}$ of user and items respectively. As a result, we have new user and item representations: $\mathbf{u} = concat(u^{(m)}, u^{(a)})$ and $\mathbf{i} = concat(i^{(m)}, i^{(a)})$. The rest of BERT-ITE-Si is similar to BERT-ITE.

In BERT-ITE, the user representation has been improved by combining information about long- and short-term preferences. As for BERT-ITE-Si, the information about these two types is enhanced. As mentioned in the subsection~\ref{subsec:notation} above, the user's side information is information about user engagement rate w.r.t the item categories, representing the user's long-term interests. And the item's side information is about the categories that it belongs to. Then the BERT-based module will make use of information from these items for user's short-term interests. So in this way, the user long- and short-term preferences have been specifically exploited and brought in more meaningful information. 


\subsection{Learning the model}
The same objective function is applied to all of our models:
\begin{equation}
\begin{aligned}
    \mathcal{L} = \eta \mathcal{L}_{I}(\hat{x}_{ui}, x_{ui}) + \mathcal{L}_{E}(\hat{y}_{ui}, y_{ui}) + \lambda \mathcal{R}(u, i)
\end{aligned}\label{eq:objective}
\end{equation}

where $\mathcal{L}_{I}$, $\mathcal{L}_{E}$ represent the objective function of implicit module and explicit module respectively, $\eta$ is a hyper-parameter to balance the effect of implicit behaviour on explicit term. $\mathcal{R}(u, i)$ is the \textit{regularization} to avoid over-fitting. 

We assume that each observable behaviour would take $0$ or $1$ as their values. For explicit data such as the ratings in n-star scale, we convert to implicit data by marking entry as 1 or 0 indicating the user has rated the item. Define:
\begin{equation}
        \begin{aligned}
            &\mathcal{L}_{I} = \sum_{(u,i)\in \mathcal{X^+} \cup \mathcal{X^-}} x_{ui}\log\hat{x}_{ui} + 
            (1 - x_{ui}) \log(1-\hat{x}_{ui}), \\
            &\mathcal{L}_{E} = \sum_{(u,i)\in \mathcal{Y^+} \cup \mathcal{Y^-}} y_{ui}\log\hat{y}_{ui} + 
            (1 - y_{ui}) \log(1-\hat{y}_{ui})
        \end{aligned}
\end{equation}

where $\mathcal{X}^+$, $\mathcal{Y}^+$ denotes the set of observed interactions in behaviour matrices $X$, $Y$ respectively. Let $\mathcal{X}^-$ denote negative instances sampled from the unobserved implicit behaviours in $X$, and $\mathcal{Y}^-$ be negative instances from the explicit matrix $Y$. The regularization $\mathcal{R}(u, i)$ is computed as:
\begin{equation}
        \begin{aligned}
            \mathcal{R}(u, i) &= \sum_{u}||\mathbf{\mathbf{p}_u}||_2^2 + \sum_{i}||\mathbf{\mathbf{q}_i}||_2^2
        \end{aligned}
 \end{equation}
 
All the models in the family of ITE use Equation \ref{eq:objective} to train the model. For optimizing this objective function, we adopt the Adam method~\cite{DBLP:journals/corr/KingmaB14}.

In particular, for models that use BERT-based module, considering the sessions where only $m \leq n$ items have interacted, we will pad into item sequence with a random number of $(n-m+1)$ items to ensure the same input size for the BERT-based module. For one pair of $(u,i) \in X^+$, if the number of items before item $i$ in the user $u$'s interaction sequence is less than $n$, the item sequence will also be padded to match the size of $n$.

\subsection{Discussion}
In this section, we discuss some advantages of our proposed models.

First, our models exploit the order of the implicit and explicit behaviours that the user makes on an item. This idea stems from the assumption that each time a user interacts with a product, the implicit behaviours appear first and logically they determine the next explicit behaviours that he/she makes on that item. For example, a user searches and clicks on a certain item, there is a high chance that he or she is interested in that item and the next action may be added to cart, then order or maybe purchase. In contrast, if a product is suggested to a user many times and is still not clicked, even hidden, we can guess that user is not interested in the product, and the continuation of suggestion can be annoying. Before us, there was NMTR \cite{DBLP:conf/icde/Gao0GCFLCJ19} which was a model that tries to exploit the ordering of the behaviours on an items. However, the association between different types of behaviour is merely the combination of the outputs as the scalar number. We consider it too simple to represent the complex relationship between behaviours. In this work, we use a deep network architecture. We believe that we do better than NMRT as well as previous models at this point.

Second, it is clear that user's preferences at any time not only depend on all items interacted in history, but also depend on the sequence of items he/she has recently interacted with. In other words, user's preferences are the mix of long- and short-term preferences. About the role of short-term preferences, for example, if someone in a certain period of time regularly clicks on items such as tables, chairs, cabinets, shelves, clocks, floor mats, etc., we can somewhat guess. Given that he has a new home and is looking to buy more items to decorate his house, his concern is the furniture at the moment. At this point, exploiting the information of these products makes sense in determining user's interests. To do this, we use the BERT-based architecture. The input of this module is the embedding vectors corresponding to user and items during a session, where the target item is the last interacted item in that session (which we want to predict if it is interacted or not). BERT module helps to learn the context information, which means we get information about the user's relationship with the item sequence with which the user has been interacting. The user embedding vector represents the user's long-term preference, the item embedding vectors represents the user's short-term preference. This means when learning the relationship between a user's short-term and long-term preferences, vector $u_{rep}$ will carry the information about both user's long- and short-term preferences. Then we can better capture user representation.

Third, as mentioned before, pure CF models learn user representation based on utility matrix only, where original representations in the form of one-hot vectors. The one-hot vectors is sometimes too simple and informative to describe the characteristics of the user as well as the item, so capturing user preferences is greatly restricted. Therefore, we try to use side information in order to more effectively describe user preferences, by enriching  long- and short-term preferences of the user. We incorporated side information into the user and item representation. Side information of user is the encoding that represents information about user's interaction rate for different categories. Throughout history, this information has been stable and reflected user's long-term preferences. In addition, adding side information for item based on the item's category information (which categories the item belongs to) also enriches item representation. On one hand, this shows consistency and match with user's side information. On other hand, in BERT-based models, that item representation also carries information about user short-term preferences. This combination is a response that improves model performance significantly. And we call this a harmonious combination.


\section{Experimental evaluation}
\label{sec:experiment}
In this section, we present our experimental settings and results. Our experiments are designed to answer the following research questions:
\begin{itemize}
    \item Does the combination of implicit and explicit behaviours
in our models work better than the state-of-the-art models, and how can the exploitation of user’s long- and short-term preferences further boost the performance?
    \item What is the influence of the different components in our models' architectures?
    \item What is the effect of hyper-parameters in the our method?
\end{itemize}

\subsection{Datasets} 
We conduct extensive experiments on two datasets: Retail Rocket and Recobell. Some statistics of Retail Rocket and Recobell are shown in Table~\ref{tab:datasets}. The details of these datasets are as follows:

\begin{table}[pos=h!]
    \centering
    \caption{Statistics of the Retailrocket and Recobell datasets}
    \begin{tabular}{|l|c|c|}
    	\hline
    		 \textbf{Dataset} & RetailRocket & Recobell \\ 
        \hline
    	\hline
    		  \textbf{Implicit \#} & 396\,923 & 2\,285\,261 \\
    		  \textbf{Explicit \#} & 18\,450 & 52\,786 \\
    		  \textbf{users \#} & 36\,751 & 206\,203 \\
    		  \textbf{items \#} & 83\,274 & 118\,293 \\
    		  \textbf{labels \#} & 1\,699 & 1\,939 \\
    		  \textbf{Sparsity} & 99.987\% & 99.999\% \\
    	\hline
    	         
    \end{tabular}
    
    \label{tab:datasets}
\end{table}

\begin{itemize}
    \item \textbf{Retail Rocket}: An e-commerce dataset taken from Kaggle\footnote{\url{https://www.kaggle.com/retailrocket/ecommerce-dataset}}. This dataset contains the information about historical interactions of users, the attributes of items and the item categories collected over a period of 4.5 months. The behaviour data includes \emph{click}, \emph{add to cart} and \emph{transaction}. In our experiments, \emph{click} is considered as implicit behaviour while \emph{add to cart} and \emph{transaction} are explicit. The users with less than 5 interactions are removed from the dataset. 
    \item \textbf{Recobell}: A dataset collected from an a-commerce website\footnote{\url{http://www.recobell.co.kr/rb/main.php?menu=pakdd2017}} in 2 months (from August 2016 to October 2016). There are 2 types of behaviours in this dataset: \emph{view} and \emph{order}, concerned as implicit behaviour and explicit behaviour respectively. Similar to Retail Rocket, the users with less than 5 interactions would be removed.
\end{itemize} 

\subsection{Evaluation scenario and metrics}

\textbf{Evaluation scenario:} We use the leave-one-out (\cite{DBLP:conf/uai/RendleFGS09, DBLP:conf/sigir/HeZKC16}) strategy to evaluate the performance. The last item in the interaction sequence of a user that received an explicit interaction from the user is used as the test item (ground-truth item), and the remaining items are left for training phase. For evaluation, we pair the ground-truth item with 999 random items (negative items) that have not been interacted with by the user. Hence, the problem becomes to rank these 1000 items and examine how the ground-truth item is ranked.

\textbf{Evaluation metrics:} We adopt two common metrics, \emph{Hit Ratio} (HR) \cite{DBLP:journals/tois/DeshpandeK04} and \emph{Normalized Discounted Cumulative Gain} (NDCG) \cite{DBLP:conf/cikm/HeCKC15}, to evaluate recommendation performance. The goal is to compute $HR@K$ and $NDCG@K$ at $K = 10$. Specifically, we compute HR@10 and NDCG@10 for each user, and the final HR@10 and NDCG@10 are averaged on the HR@10 and NDCG@10 of all users.

\textit{\textbf{Hit Ratio (HR):}} For each user, $HR@K$ corresponds to whether the ground-truth item appears in the top K items ranked for that user.
\begin{equation}
    HR@K =
  \begin{cases}
       1, & \text{if the ground-truth item is in top K,} \\
       0, & \text{otherwise.} \\
  \end{cases}
\end{equation}

\textit{\textbf{Normalized Discounted Cumulative Gain (NDCG):}} Ra- ther than checking the existence of the ground-truth item in the top K as \emph{Hit Ratio}, $NDCG@K$ takes into consideration the order of the ground-truth item in the top K items. The higher the $NDCG@K$ score is, the better the ranking is. $NDCG@K$ score for each user is as follows:

\begin{equation}
  NDCG@K=\begin{cases}
    \frac{log(2)}{log(i+1)},&\text{if the ground-truth item is }\\ 
    &\text{ranked at position $i$},\\
    0,&\text{otherwise}.
  \end{cases}
\end{equation}

\subsection{Baselines and Implementation Details}

\begin{figure*}[pos=!ht]
    \centering
    \captionsetup{justification=centering}
    \includegraphics[width=16cm, height=9cm]{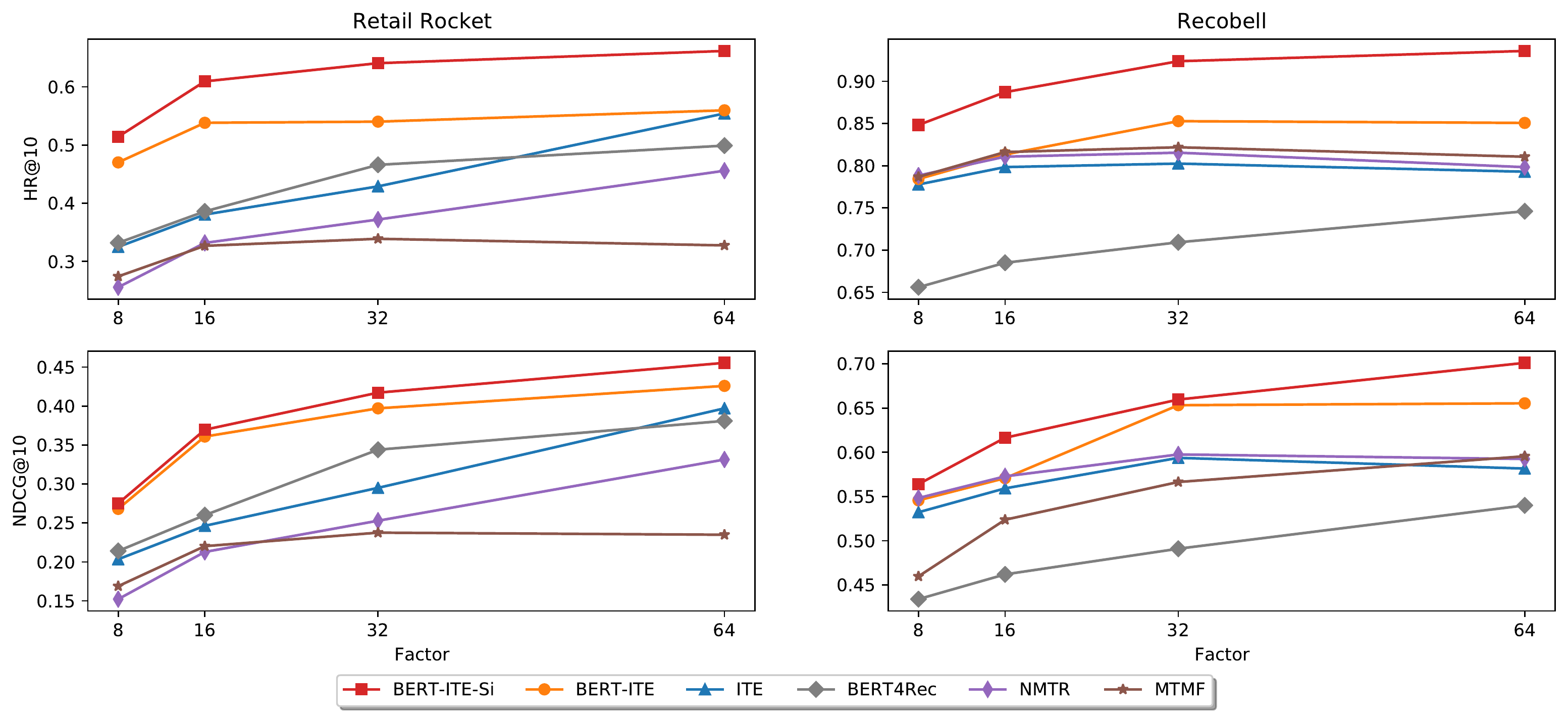}
    \caption{Comparison of main models and baselines when the number of factors varies}
    \label{fig-k-vary-main}
\end{figure*}

\begin{figure*}[pos=!ht]
    \centering
    \captionsetup{justification=centering}
    \includegraphics[width=16cm, height=9cm]{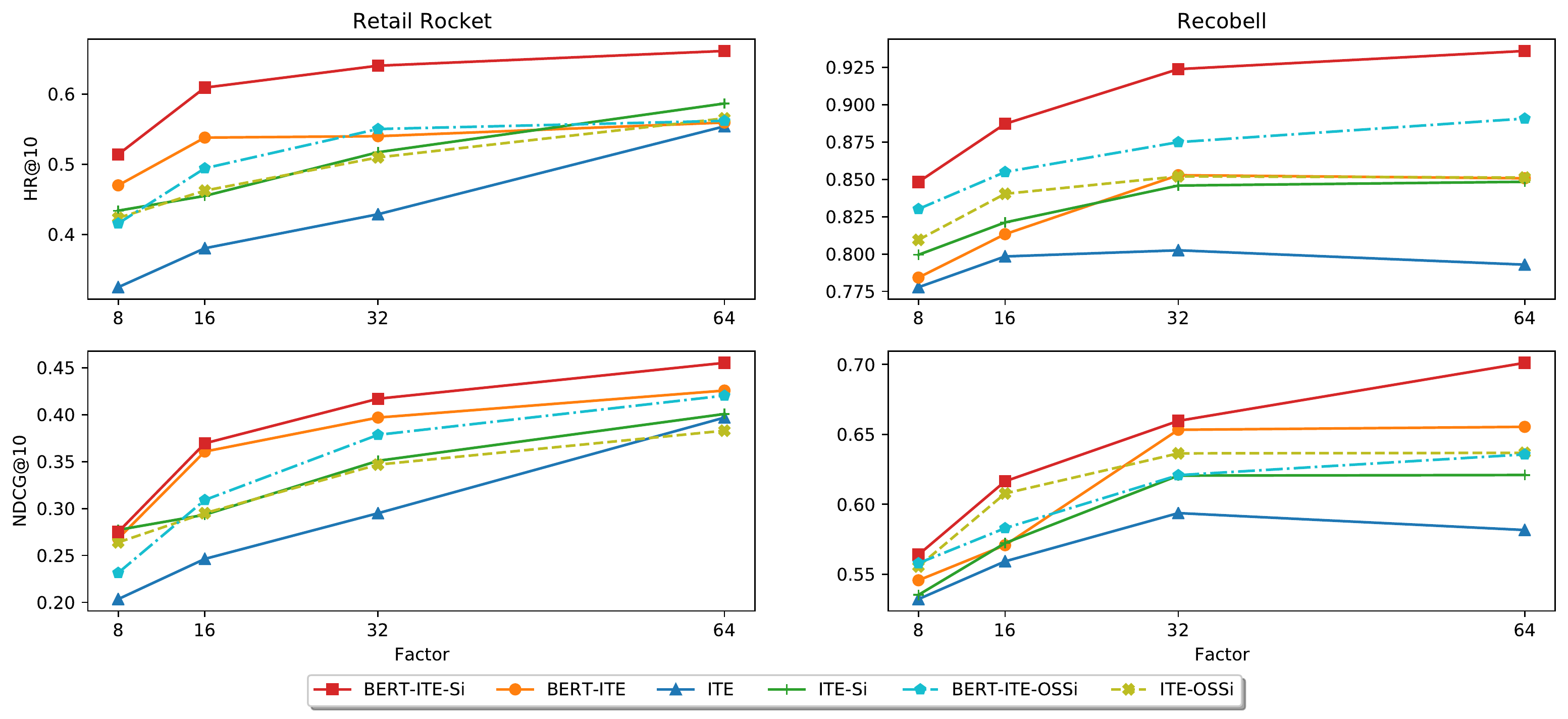}
    \caption{Comparison of main models and variants when the number of factors varies}
    \label{fig-k-vary-ablation}
\end{figure*}

\begin{figure}[pos=!ht]
    \centering
    \captionsetup{justification=centering}
    \includegraphics[scale=0.5]{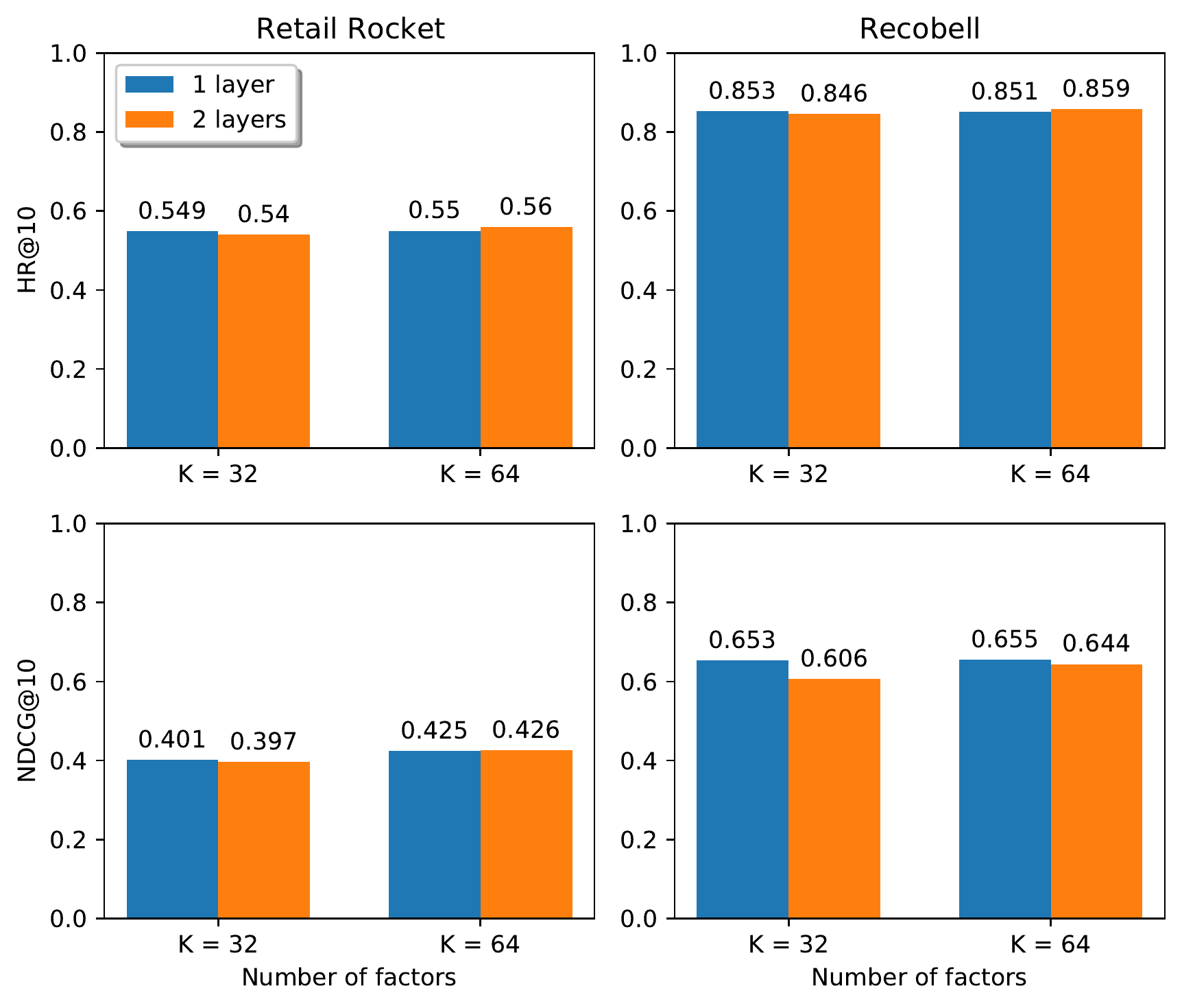}
    \caption{Performance of BERT-ITE when the number of self-attention layers varies}
    \label{fig-num-block}
\end{figure}

\begin{figure}[pos=!ht]
    \centering
    \captionsetup{justification=centering}
    \includegraphics[scale=0.55]{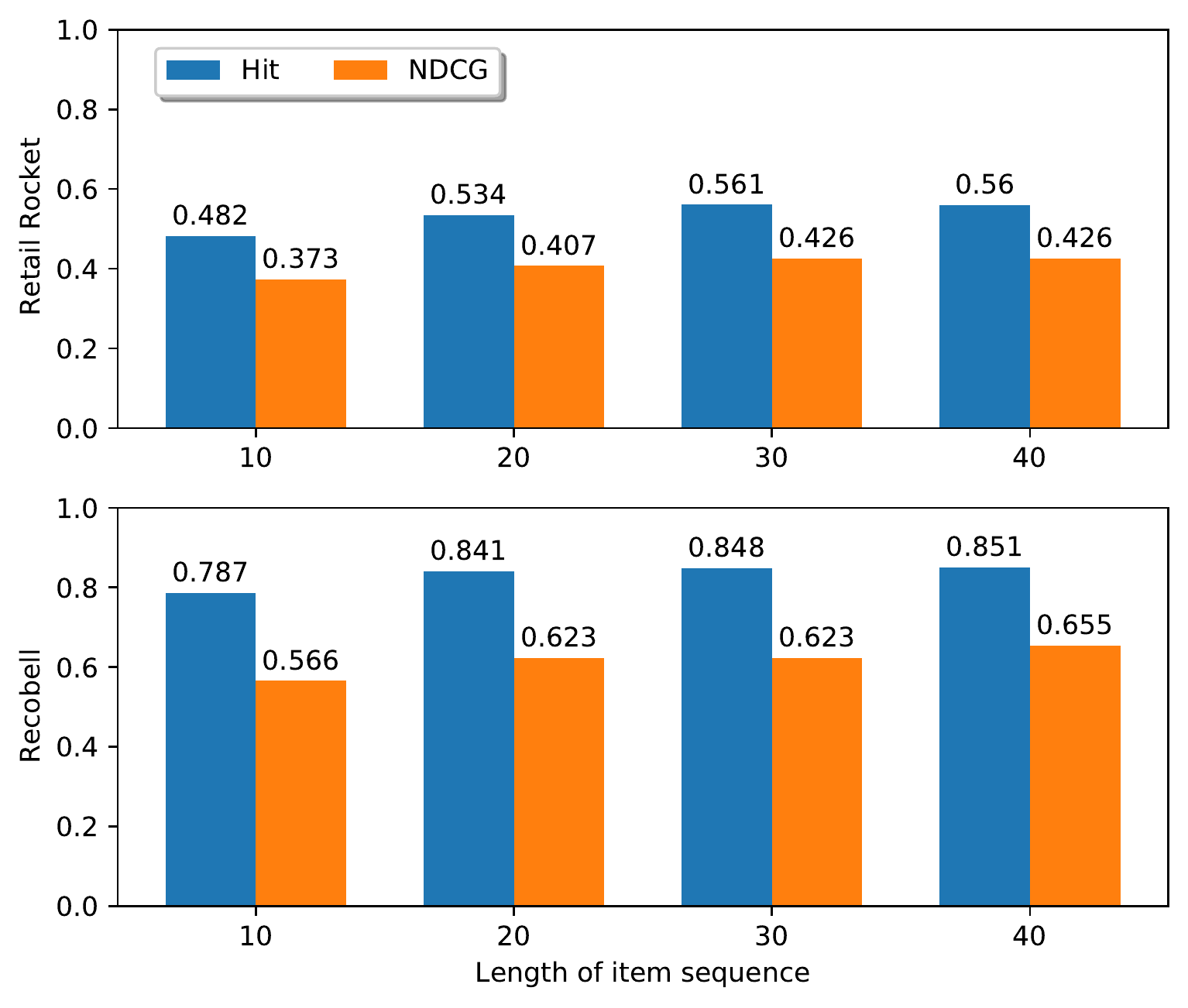}
    \caption{Performance of BERT-ITE with different lengths of item sequence}
    \label{fig-num-seq-vary}
\end{figure}



\textbf{Model in use:} We evaluate the models that have been discussed in Section~\ref{sec:pro_model}, which are \textbf{ITE}, \textbf{BERT-ITE} and \textbf{BE- RT-ITE-Si}, to show the improvement in recommendation performance when combining implicit and explicit behaviours complexly, and how user's long- and short-term preferences can further boost the performance. Additionally, we experiment with three other models. Two of them are the variants of \textbf{ITE}, named as \textbf{ITE-Si} and \textbf{ITE-OSSi}. \textbf{ITE-Si} is expanded similarly to \textbf{BERT-ITE-Si}, that incorporates side information into both user and item representations. Meanwhile, \textbf{ITE-OSSi} only integrates side information into item representation. Particularly, in \textbf{ITE-OSSi}, only $i^{(m)}$ is concatenated with $i^{(a)}$. The last model is \textbf{BERT-ITE-OSSi}, which behaves similarly to \textbf{ITE-OSSi}, that only incorporates side information into item representation. We add \textbf{ITE-Si} to examine how \textbf{ITE} can capture user's long-term preference encoded as side information, and the two \textbf{-OSSi} models to solely analyze the impact of category on the representation of item. We refer \textbf{ITE}, \textbf{BERT-ITE}, \textbf{BERT-ITE-Si} as main models, and the others as variants.

\textbf{Baselines:} To verify the effectiveness of our method, we compare our proposed models with the following baselines in which two of them consider both implicit and explicit feedback and the other considers user's item sequence.
\begin{itemize}
    \item \textbf{NMTR \cite{DBLP:conf/icde/Gao0GCFLCJ19}}: A model that learns the cascading relationship among multiple types of behaviours in a multi-task learning framework.
    \item \textbf{MTMF \cite{DBLP:conf/cscwd/ShiLLZG17}}: A model that separately considers multiple types of interactions (\emph{view}, \emph{want} and \emph{click}) to enhance the quality of rating prediction. 
    \item \textbf{BERT4Rec \footnote{\url{https://github.com/jaywonchung/BERT4Rec-VAE-Pytorch}} \cite{DBLP:conf/cikm/SunLWPLOJ19}}: A model that models user's sequence of interacted items bidirectionally using Cloze task to predict the masked items. 
\end{itemize} 

Note that \textbf{BERT4Rec} only considers one type of behaviour, thus we experiment it with only implicit feedback.

\textbf{Parameter settings:} Some common hyper-parameters have great impact on all the models: $lr$ (learning rate), \textit{batch size}, $\eta$ (the parameter that controls the influence of implicit loss), and $K$, or \emph{num factors} (the dimension of the implicit vector). We tune the hyper-parameters with \textit{batch size} $\in$ \{512, 1024, 2048, 4096\}, $lr$ $\in$ \{0.0001, 0.0005, 0.001, 0.002\}, $\eta$ $\in$ \{0.1, 0.5, 1.0, 2.0\}. The result of this tuning process gives us $lr$ = 0.001 for all the model; \emph{batch size} = 512 for BERT-ITE model and its variants, 2048 for all other models; $\eta$ = 1.0 for NMTR model and 0.5 for the remaining models. To investigate the impact of K on the performance, we conduct experiments to compare the models with K $\in$ [8, 16, 32, 64] while other hyper-parameters are fixed. It is worth to note that the models might face over-fitting if K is too large \cite{DBLP:conf/www/HeLZNHC17}. Furthermore, we examine the influence of the number of Transformer layers and $\eta$ on our models on both datasets, and the behaviour of models when gradually increasing the number of epochs on the Retail Rocket.  


As mentioned above, the default versions of BERT-ITE and its variants have two Transformer layers which in turn includes a two-head self-attention. For ITE, ITE-Si, ITE-OSSi and NMTR which adopt the architecture of NeuMF \cite{DBLP:conf/www/HeLZNHC17}. Note that $K$ in these model is the dimension size of the implicit vector $\phi^I$. In BERT-ITE and its variants, $K$ could also be referred to the same thing as originally $K$ is the dimension size of user/item embedding and this value remained unchanged after the BERT-based module. For all the models, 9 negative items are sampled per one positive item when computing the loss. 


\subsection{Performance comparison}
We will show the performance comparison among our proposed models and baselines with different values of K on the two datasets. Moreover, we will investigate the impact of some components on our proposed models: the number of Transformer layers \emph{L}, the length of item sequence and $\eta$. Additionally, we will present the change in performance of models when the number of epoch gradually increases on Retail Rocket. The figures reported in these experiments except the last one are the highest scores achieved in the test set.

\begin{figure*}[pos=!ht]
    \centering
    \captionsetup{justification=centering}
    \includegraphics[scale=0.65]{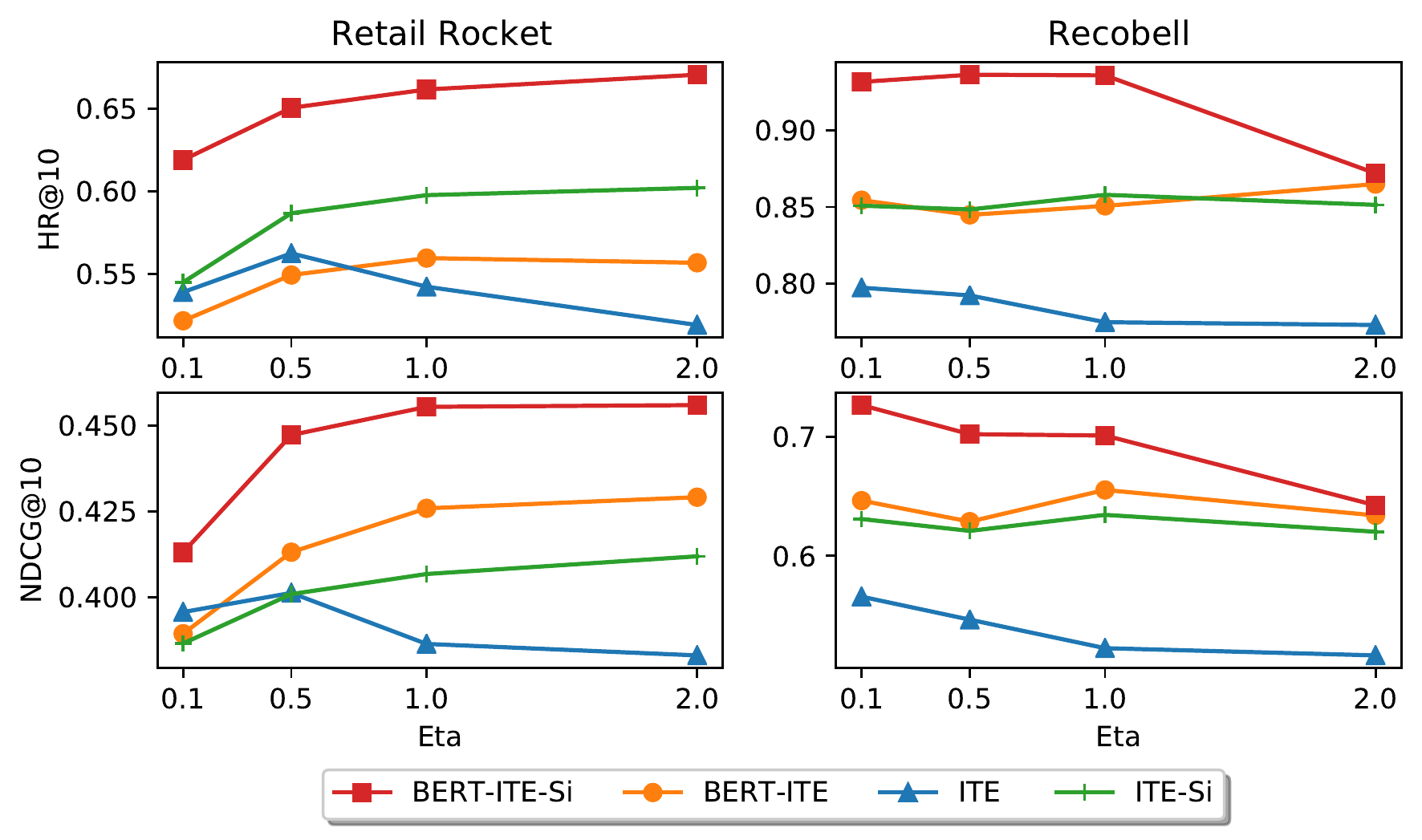}
    \caption{Performance of BERT-ITE, BERT-ITE-Si, ITE, ITE-Si when eta varies}
    \label{fig-eta-vary}
\end{figure*}

\subsubsection{Performance comparison with different values of K}

\begin{figure*}[pos=!ht]
    \centering
    \captionsetup{justification=centering}
    \includegraphics[width=\textwidth]{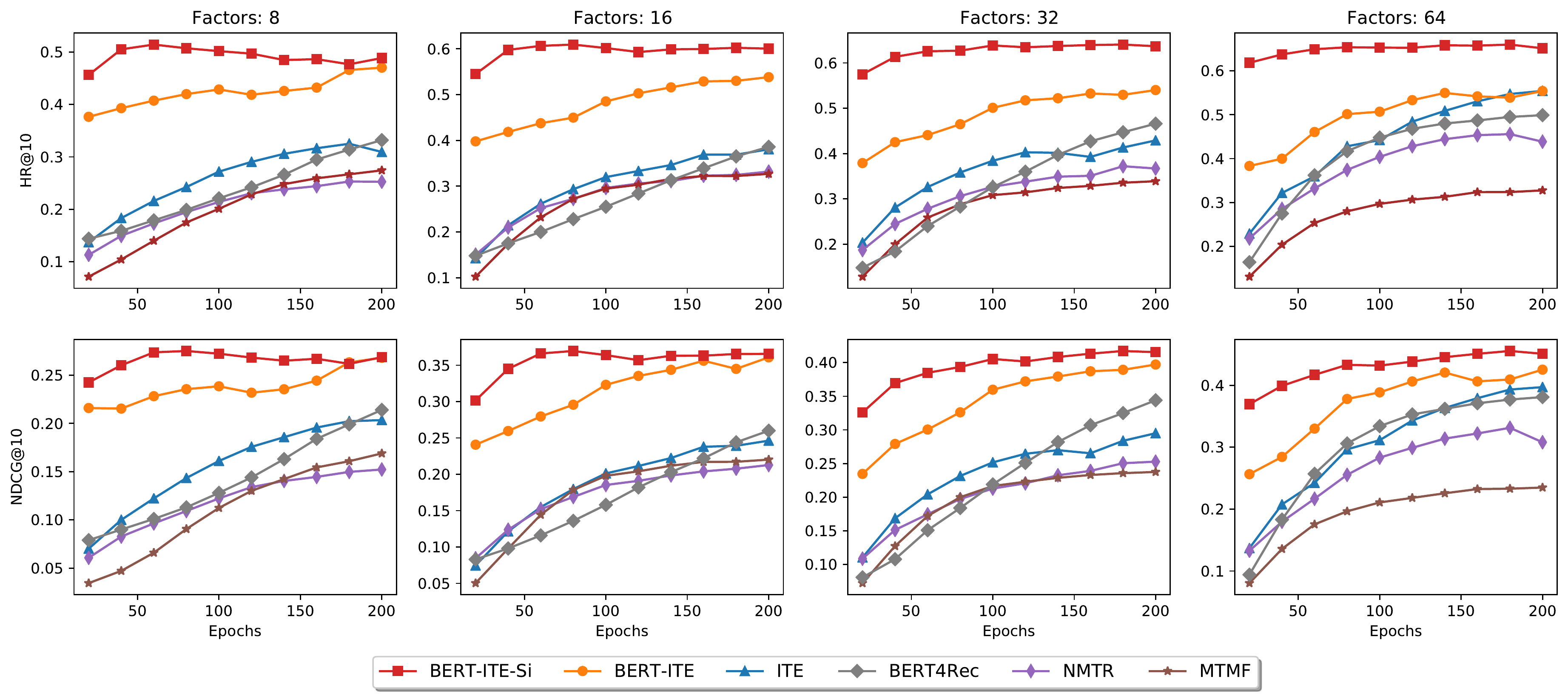}
    \caption{Comparison of the  various models in \textbf{Retail Rocket} when increasing the number of epochs gradually. From left to right: $K = 8, 16, 32, 64$}
    \label{fig-epoch-k-vary-main}
\end{figure*}

\begin{figure*}[pos=!ht]
    \centering
    \captionsetup{justification=centering}
    \includegraphics[width=\textwidth]{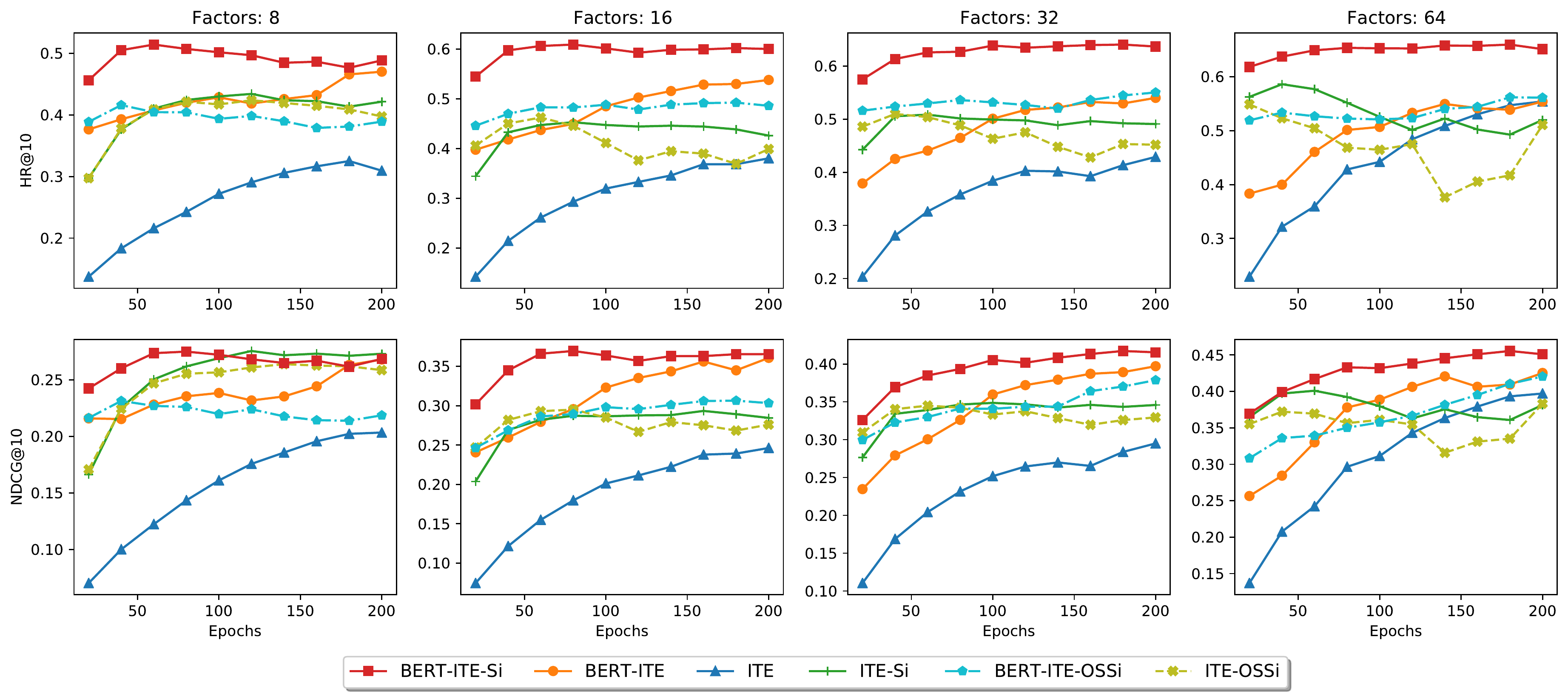}
    \caption{Comparison of the various models in \textbf{Retail Rocket} when increasing the number of epochs gradually. From left to right: $K = 8, 16, 32, 64$}
    \label{fig-epoch-k-vary-ablation}
\end{figure*}

\begin{figure*}[pos=!ht]
    \centering
    \captionsetup{justification=centering}
    \includegraphics[width=0.9\textwidth]{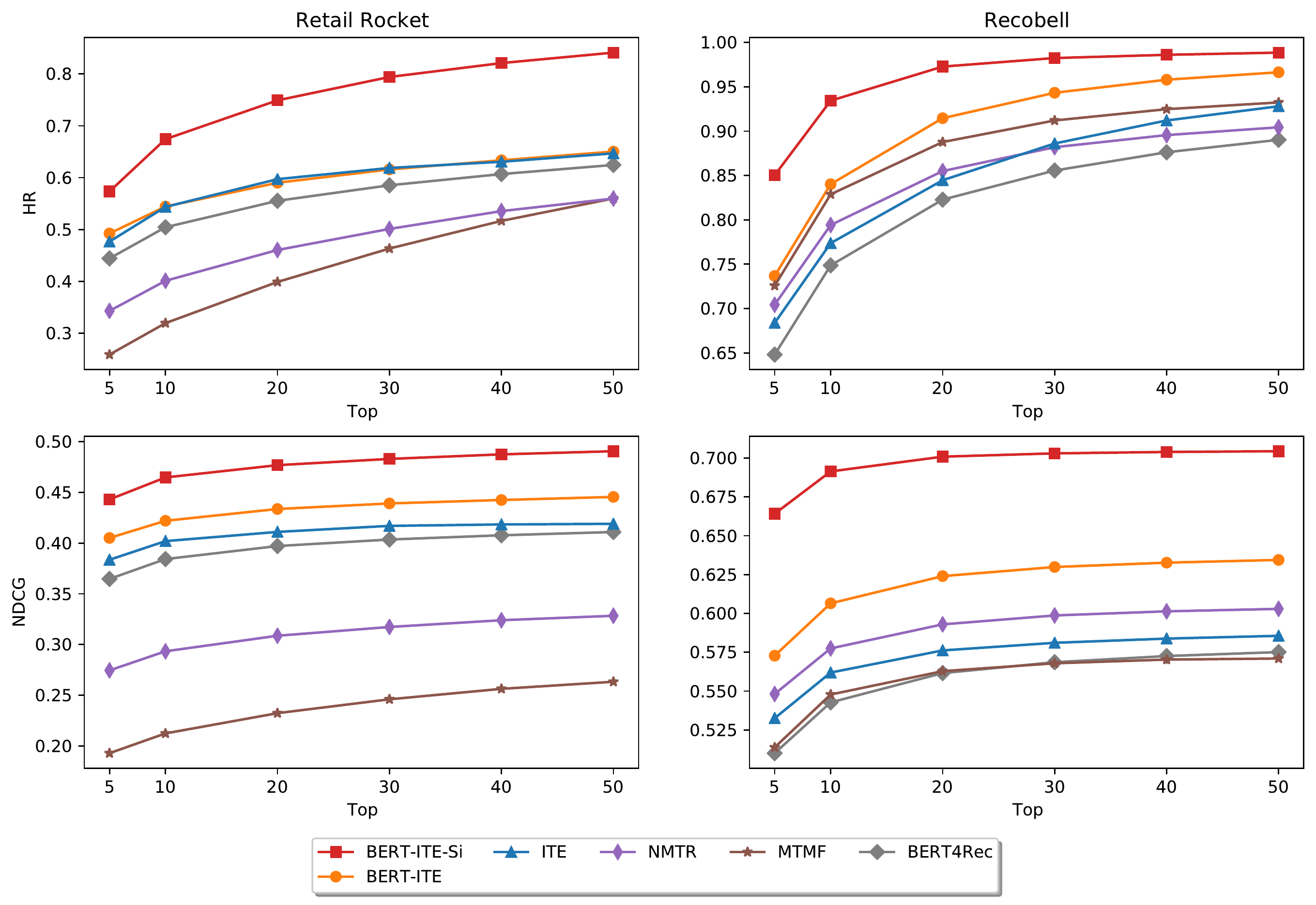}
    \caption{Comparison of main models and baselines when Top @K varies}
    \label{fig-topk-ablation}
\end{figure*}

Fig.~\ref{fig-k-vary-main} shows the best HR@10 and NDCG@10 scores of our main models (ITE, BERT-ITE, BERT-ITE-Si) and baselines with respect to \textbf{the num factors} $K \in [8, 16, 32, 64]$ on two datasets. The general trend is that both scores are higher when K increases. The figure shows that on all values of K, ITE achieves higher results than MTMF and NMTR on Retail Rocket and comparable results to those on Recobell, which proves our efficient exploitation of the order from implicit to explicit feedback. The performance is further enhanced when the recent item sequence is taken into consideration. This is shown by the outperformance of BERT-ITE over ITE for both metrics. For example, on Retail Rocket at K = 8, where ITE, MTMF, NMTR perform worst among four values of K, it achieves rather good scores. Therefore, it can be drawn that recent item sequence was effectively exploited to enrich user representation, which is absent in the other three models. When comparing BERT-ITE and BERT4Rec, two methods that both model item sequence, BERT-ITE is still superior. The reason for this could be BERT-ITE uses both explicit and implicit feedback on one item, while BERT4Rec only considers implicit type. This confirms the usefulness of the information about multiple types of feedback. Furthermore, when user and item one-hot embeddings are incorporated with information about item category, the performances of BERT-ITE is boosted significantly. Particularly, the red line (BERT-ITE-Si) always lies on the top of each sub-figure. This is the evidence for the benefit of side information in enhancing user's long- and short-term preferences. 

Fig.~\ref{fig-k-vary-ablation} depicts the comparison of ITE, BERT-ITE and their side-information variants. Firstly, the value of user's long-term preference encoded as item category is once again verified when ITE-Si is superior to ITE. Secondly, ITE-OSSI, the result of solely using item category information on item encodings, boosts the performance of ITE. It even achieves equivalent results compared to ITE-Si. If we compare the improvement in performance of ITE-Si compared to ITE-OSSi and BERT-ITE-SI compared to BERT-ITE-OSSi, it can be seen that long-term information does not contribute to ITE-OSSi as much as to BERT-ITE-OSSi. This might be due to the self-attention mechanism in BERT-ITE which could learn the impact of long-term information more effectively than NeuMF in ITE.

\subsubsection{Ablation Study} 
\textbf{Impact of the number of Transformer layers on BERT-ITE}

Fig.~\ref{fig-num-block} shows the performance of BERT-ITE model with the number of Transformer layers is set to 1 and 2, with $K \in \{32, 64\}$. As can be seen, for both datasets and for each value of K, both HR@10 and NDCG@10 scores are nearly the same as the number of self-attention layers is 1 or 2 . This might be due to the sparsity of the two datasets that using one layer already can learn how much each item in the item sequence should be focused on.

\textbf{Impact of the length of item sequence on BERT-ITE}

Fig.~\ref{fig-num-seq-vary} show the changes in performance of BERT-ITE when considering 10, 20, 30 and 40 most recent items in user history to learn his/her short-term preference. For both datasets, the results slightly increase as more items are considered. Using only 10 items seems insufficient to capture the recent interest of a user. As suggested from the figure, this number should be from 20 to 40 items to achieve decent results.

\textbf{Impact of $\eta$ on ITE and BERT-ITE}

Fig.~\ref{fig-eta-vary} shows the change in the performance of four models: ITE, ITE-SI, BERT-ITE and BERT-ITE-Si for some given values of $\eta$. $\eta$ is a hyper-parameter on $\mathcal{L}_{I}$ of loss function $\mathcal{L}$ to consider the correlating role of implicit and explicit module. 

On Retail Rocket, in general, all three models: BERT-ITE-Si, BERT-ITE and ITE-Si tend to increase slightly according to the value of $\eta$. These model have the scores of HR@10 and NDCG@10 when $\eta=1$ is approximate to when $\eta=2$. Meanwhile, ITE has the best performance when $\eta=0.5$ and drops quite a lot at $\eta=1$ or $\eta=2$. On Recobell, BERT-ITE and ITE-Si have rather stable results and the best results are all achieved at $\eta=1$. The performance of BERT-ITE-Si does not change too much when $\eta \in \{0.1, 0.5, 1\}$ and decreases sharply at $\eta=2$. ITE is the best when $\eta=0.1$ instead of $0.5$ as on Retail Rocket. Thus, we find that in most cases the implicit module plays a more important role than the explicit module in BERT-based architecture and ITE-Si, which is opposite to ITE. Note that to remain consistency, we still choose $\eta=0.5$ for ITE and ITE-Si to conduct other experiments on RetailRocket and Recobell since the scores at this value is only slightly worse than the best ones.

\textbf{Model comparison with number of epochs gradually increases during training}

Fig.~\ref{fig-epoch-k-vary-main} and Fig.\ref{fig-epoch-k-vary-ablation} show performance of all models on Retail Rocket as the number of epochs increases. Each epoch is a learning cycle in which the model learns through all the training data. We can see that, in most cases, our proposed models outperform the baselines. In particular, BERT-ITE-Si always outstrips the remaining models. Additionally, models with side-information tend to converge sooner than the others. This saves us training time to get a good performance model. 

\textbf{Model comparison with the change of top @K}

We have conducted some more experiments to compare the performances of methods in terms of of NDCG and HR when changing top @K on Retail Rocket and Recobell datasets. The experimental results are shown in Figure \ref{fig-topk-ablation}. It is clear that NDCG and HR scores of all methods increase when top @K increases and Figure \ref{fig-topk-ablation} illustrates this property.

Overall, the performance rankings of these methods almost do not change as top @K increases. In addition, BERT-ITE-Si and BERT-ITE are always on top. On Retail Rocket dataset, the orders of the methods in terms of HR and NDCG are stable. All of our proposed methods (BERT-ITE-Si, BERT-ITE and ITE) outperform the baselines: MTMF, NMTR and BERT4Rec. BERT-ITE is comparable to ITE in terms of HR but is superior to ITE on NDCG. Especially, exploiting side information (BERT-ITE-Si) improves the performances of BERT-ITE with significant magnitudes. On Recobell dataset, except for ITE, our proposed methods achieve higher results than the baselines. At first, the HR score of ITE is smaller than NMTR's and MTMF's, then it surpasses NMTR's as well as approaches MTMF's when top@K increases. Regarding NDCG score, ITE is only inferior to NMTR but is superior to MTMF and BERT4Rec.

\section{Conclusion}
\label{sec:conclusion}

We have proposed ITE, BERT-ITE and BERT-ITE-Si, the neural networks for modeling the sequence of user behaviours in real-world. Our models successfully learn the complex relation between implicit and explicit behaviours, thus make the prediction more precisely. In   particular, our BERT-ITE and BERT-ITE-Si models help to better capture user interests by harmoniously combining information about long- and short-term preferences. Especially when side information is utilized in BERT-ITE-Si, the performance of the model is greatly improved. Apart from improving the model performance, the extended version BERT-ITE-Si is expected to avoid the cold start thanks to additional information attached in the input layer. We left further study on these benefits for future work. In addition, extensive experiments show that our models outperform the state-of-the-art methods. Moreover, BERT-ITE-Si gains higher experimental results than the original version in the two e-commerce datasets, which indicates the effectiveness of side information.


\bibliographystyle{cas-model2-names}
\bibliography{bert_ite}

\end{document}